# Learning class-to-class selectional preferences

**Eneko Agirre**
IXA NLP Group
University of the Basque Country
649 pk. 20.080
Donostia. Spain.
`eneko@si.ehu.es`

**David Martinez**
IXA NLP Group
University of the Basque Country
649 pk. 20.080
Donostia. Spain.
`jibmaird@si.ehu.es`

## Abstract

Selectional preference learning methods have usually focused on word-to-class relations, e.g., a verb selects as its subject a given nominal class. This papers extends previous statistical models to class-to-class preferences, and presents a model that learns selectional preferences for classes of verbs. The motivation is twofold: different senses of a verb may have different preferences, and some classes of verbs can share preferences. The model is tested on a word sense disambiguation task which uses subject-verb and object-verb relationships extracted from a small sense-disambiguated corpus.

## 1 Introduction

Previous literature on selectional preference has usually learned preferences for words in the form of classes, e.g., the object of eat is an edible entity. This paper extends previous statistical models to classes of verbs, yielding a relation between classes in a hierarchy, as opposed to a relation between a word and a class.

The model is trained using subject-verb and object-verb associations extracted from Semcor, a corpus (Miller et al., 1993) tagged with WordNet word-senses (Miller et al., 1990). The syntactic relations were extracted using the Minipar parser (Lin, 1993). A peculiarity of this exercise is the use of a small sense-disambiguated corpus, in contrast to using a large corpus of ambiguous words. We think that two factors can help alleviate the scarcity of data: the fact that using disambiguated words provides purer data, and the ability to use classes of verbs in the preferences. Nevertheless, the approach can be easily extended to larger, non-disambiguated corpora.

We have defined a word sense disambiguation exercise in order to evaluate the extracted preferences, using a sample of words and a sample of documents, both from Semcor.

Following this short introduction, section 2 reviews selectional restriction acquisition. Section 3 explains our approach, which is formalized in sections 4 and 5. Next, section 6 shows the results on the WSD experiment. Some of the acquired preferences are analysed in section 7. Finally, some conclusions are drawn and future work is outlined.

## 2 Selectional preference learning

Selectional preferences try to capture the fact that linguistic elements prefer arguments of a certain semantic class, e.g. a verb like '*eat*' prefers as object edible things, and as subject animate entities, as in, (1) "*She was eating an apple*". Selectional preferences get more complex than it might seem: (2) "*The acid ate the metal*", (3) "*This car eats a lot of gas*", (4) "*We ate our savings*", etc.

Corpus-based approaches for selectional preference learning extract a number of (e.g. verb/subject) relations from large corpora and use an algorithm to generalize from the set of nouns for each verb separately. Usually, nouns are generalized using classes (concepts) from a lexical knowledge base (e.g. WordNet).

Resnik (1992, 1997) defines an information-theoretic measure of the association between a verb and nominal WordNet classes: selectional association. He uses verb-argument pairs from

Brown. Evaluation is performed applying intuition and WSD. Our measure follows in part from his formalization.

Abe and Li (1995) follow a similar approach, but they employ a different information-theoretic measure (the minimum description length principle) to select the set of concepts in a hierarchy that generalize best the selectional preferences for a verb. The argument pairs are extracted from the WSJ corpus, and evaluation is performed using intuition and PP-attachment resolution.

Stetina et al. (1998) extract word-arg-word triples for all possible combinations, and use a measure of "relational probability" based on frequency and similarity. They provide an algorithm to disambiguate all words in a sentence. It is directly applied to WSD with good results.

## 3 Our approach

The model explored in this paper emerges as a result of the following observations:

- Distinguishing verb senses can be useful. The examples for *eat* above are taken from WordNet, and each corresponds to a different word sense[1]: example (1) is from the "*take in solid food*" sense of *eat,* (2) from the "*cause to rust*" sense, and examples (3) and (4) from the *"use up"* sense.
- If the word senses of a set of verbs are similar (e.g. word senses of ingestion verbs like *eat, devour, ingest,* etc.) they can have related selectional preferences, and we can generalize and say that a class of verbs has a particular selectional preference.

Our formalization thus distinguishes among verb senses, that is, we treat each verb sense as a different unit that has a particular selectional preference. From the selectional preferences of single verb word senses, we also infer selectional preferences for classes of verbs.

Contrary to other methods (e.g. Li and Abe's), we don't try to find the classes which generalize best the selectional preferences. All possibilities, even the very low probability ones, are stored.

The method stands as follows: we collect [*noun-word-sense relation verb-word-sense*] triples from Semcor, where the relation is either subject or object. As word senses refer to concepts, we also collect the triple for each possible combination of concepts that subsume the word senses in the triple. Direct frequencies and estimates of frequencies for classes are then used to compute probabilities for the triples.

These probabilities could be used to disambiguate either nouns, verbs or both at the same time. For the time being, we have chosen to disambiguate nouns only, and therefore we compute the probability for a nominal concept, given that it is the subject/object of a particular verb. Note that when disambiguating we ignore the particular sense in which the governing verb occurs.

## 4 Formalization

As mentioned in the previous sections we are interested in modelling the probability of a nominal concept given that it is the subject/object of a particular verb:

$$P(cn_i \mid rel\ v) \qquad (1)$$

Before providing the formalization for our approach we present a model based on words and a model based on nominal-classes. Our class-to-class model is an extension of the second[2]. The estimation of the frequencies of classes are presented in the following section.

---

[1] A note is in order to introduce the terminology used in the paper. We use concept and class indistinguishably, and they refer to the so-called *synsets* in WordNet. Concepts in WordNet are represented as sets of synonyms, e.g. <*food, nutrient*>. A word sense in WordNet is a word-concept pairing, e.g. given the concepts *a*=<*chicken, poulet, volaille*> and *b*=<*wimp, chicken, crybaby*> we can say that chicken has at least two word senses, the pair *chicken-a* and the pair *chicken-b*. In fact the former is sense 1 of chicken, and the later is sense 3 of chicken. For the sake of simplicity we also talk about <*chicken, poulet, volaille*> being a word sense of chicken.

[2] Notation: *v* stands for a verb, *cn* (*cv*) stand for nominal (verbal) concept, $cn_i$ ($cv_i$) stands for the concept linked to the *i*-th sense of the given noun (verb), *rel* could be any grammatical relation (in our case object or subject), $\subseteq$ stands for the subsumption relation, *fr* stands for frequency and $\hat{fr}$ for the estimation of the frequencies of classes.

$$P(cn_i \mid rel\,v) = \sum_{cn \supseteq cn_i} P(cn_i \mid cn) \times P(cn \mid rel\,v) = \sum_{cn \supseteq cn_i} \frac{\hat{fr}(cn_i, cn)}{\hat{fr}(cn)} \times \frac{\hat{fr}(cn\,rel\,v)}{fr(rel\,v)} \quad (3)$$

$$P(cn_i \mid rel\,v) = \max_{cv_j\, sense\, of\, v} \sum_{cn \supseteq cn_i} \sum_{cv \supseteq cv_j} P(cn_i \mid cn) \times P(cv_j \mid cv) \times P(cn \mid rel\,cv)$$

$$= \max_{cv_j\, sense\, of\, v} \sum_{cn \supseteq cn_i} \sum_{cv \supseteq cv_j} \frac{\hat{fr}(cn_i, cn)}{\hat{fr}(cn)} \times \frac{\hat{fr}(cv_j, cv)}{\hat{fr}(cv)} \times \frac{\hat{fr}(cn\,rel\,cv)}{fr(rel\,cv)} \quad (4)$$

$$\hat{fr}(cn) = \sum_{cn_i \subseteq cn} \frac{1}{classes(cn_i)} \times fr(cn_i) \quad (5)$$

$$\hat{fr}(cn_i, cn) = \begin{cases} \sum_{cn_j \subseteq cn_i} \frac{1}{classes(cn_j)} \times fr(cn_j) & \text{if } cn_i \subseteq cn \\ 0 & \text{otherwise} \end{cases} \quad (6)$$

$$\hat{fr}(cn\,rel\,v) = \sum_{cn_i \subseteq cn} \frac{1}{classes(cn_i)} \times fr(cn_i\, rel\, v) \quad (7)$$

$$\hat{fr}(cn\,rel\,cv) = \sum_{cn_i \subseteq cn} \sum_{cv_i \subseteq cn} \frac{1}{classes(cn_i)} \times \frac{1}{classes(cv_i)} \times fr(cn_i\, rel\, cv_i) \quad (8)$$

### 4.1 Word-to-word model: eat chicken$_i$

At this stage we do not use information of class subsumption. The probability of the first sense of *chicken* being an object of *eat* depends on how often does the concept linked to *chicken$_1$* appear as object of the word *eat*, divided by the number of occurrences of *eat* with an object.

$$P(cn_i \mid rel\,v) = \frac{fr(cn_i\, rel\, v)}{fr(rel\, v)} \quad (2)$$

Note that instead of $P(sense_i \mid rel\,v)$ we use $P(cn_i \mid rel\,v)$, as we count occurrences of concepts rather than word senses. This means that synonyms also count, e.g. *poulet* as synonyms of the first sense of *chicken*.

### 4.2 word-to-class model: eat <food, nutrient>

The probability of *eat chicken$_1$* depends on the probabilities of the concepts subsumed by and subsuming *chicken$_1$* being objects of eat. For instance, if *chicken$_1$* never appears as an object of eat, but other word senses under <*food, nutrient*> do, the probability of *chicken$_1$* will not be 0.

Formula (3) shows that for all concepts subsuming $cn_i$ the probability of $cn_i$ given the more general concept times the probability of the more general concept being a subject/object of the verb is added. The first probability is estimated dividing the class frequencies of $cn_i$ with the class frequencies of the more general concept. The second probability is estimated as in 4.1.

### 4.3 class-to-class model: <ingest, take in, …> <food, nutrient>

The probability of *eat chicken$_1$* depends on the probabilities of all concepts above *chicken$_1$* being objects of all concepts above the possible senses of *eat*. For instance, if *devour* never appeared on the training corpus, the model could infer its selectional preference from that of its superclass <ingest, take in, ...>. As the verb can be polysemous, the sense with maximum probability is selected.

Formula (4) shows that the maximum probability for the possible senses ($cv_j$) of the verb is taken. For each possible verb concept ($cv$) and noun concept ($cn$) subsuming the target concepts ($cn_i, cv_j$), the probability of the target concept given the subsuming concept (this is done twice, once for the verb, once for the noun) times the probability the nominal concept being subject/object of the verbal concept is added.

## 5 Estimation of class frequencies

Frequencies for classes can be counted directly from the corpus when the class is linked to a word sense that actually appears in the corpus, written as $fr(cn_i)$. Otherwise they have to be estimated using the direct counts for all subsumed concepts, written as $\hat{fr}(cn_i)$. Formula (5) shows that all the counts for the subsumed concepts ($cn_i$) are added, but divided by the number of classes for which $c_i$ is a subclass (that is, all ancestors in the hierarchy). This is necessary to guarantee the following:

$$\sum_{cn \supseteq cn_i} P(cn_i \mid cn) = 1.$$

Formula (6) shows the estimated frequency of a concept given another concept. In the case of the first concept subsuming the second it is 0, otherwise the frequency is estimated as in (5).

Formula (7) estimates the counts for [nominal-concept relation verb] triples for all possible nominal-concepts, which is based on the counts for the triples that actually occur in the corpus. All the counts for subsumed concepts are added, divided by the number of classes in order to guarantee the following:

$$\sum_{cn} P(cn \mid subj\ v) = 1$$

Finally, formula (8) extends formula (7) to [nominal-concept relation verbal-concept] in a similar way.

## 6 Training and testing on a WSD exercise

For training we used the sense-disambiguated part of Brown, Semcor, which comprises around

| Noun | # sens | # occ | # occ. as obj | # occ. as subj |
|---|---|---|---|---|
| account | 10 | 27 | 8 | 3 |
| age | 5 | 104 | 10 | 9 |
| church | 3 | 128 | 19 | 10 |
| duty | 3 | 25 | 8 | 1 |
| head | 30 | 179 | 58 | 16 |
| interest | 7 | 140 | 31 | 13 |
| member | 5 | 74 | 13 | 11 |
| people | 4 | 282 | 41 | 83 |
| Overall | 67 | 959 | 188 | 146 |

**Table 1**. Data for the selected nouns.

| | Obj | | | Subj | | |
|---|---|---|---|---|---|---|
| | Prec. | Cov. | Rec. | Prec. | Cov. | Rec. |
| Random | .192 | 1.00 | .192 | .192 | 1.00 | .192 |
| MFS | .690 | 1.00 | .690 | .690 | 1.00 | .690 |
| Word2word | .959 | .260 | .249 | .742 | .243 | .180 |
| Word2class | .669 | .867 | .580 | .562 | .834 | .468 |
| Class2class | .666 | .973 | .648 | .540 | .995 | .537 |

**Table 2**. Average results for the 8 nouns.

250.000 words tagged with WordNet word senses. The parser we used is Minipar. For this current experiment we only extracted verb-object and verb-subject pairs. Overall 14.471 verb-object pairs and 12.242 verb-subject pairs were extracted. For the sake of efficiency, we stored all possible class-to-class relations and class frequencies at this point, as defined in formulas (5) to (8).

The acquired data was tested on a WSD exercise. The goal was to disambiguate all nouns occurring as subjects and objects, but it could be also used to disambiguate verbs. The WSD algorithm just gets the frequencies and computes the probabilities as they are needed. The word sense with the highest probability is chosen.

Two experiments were performed: on the lexical sample we selected a set of 8 nouns at random[3] and applied 10fold crossvalidation to make use of all available examples. In the case of whole documents, they were withdrawn from the training corpus and tested in turn.

[3] This set was also used on a previous paper (Agirre & Martinez, 2000).

|       | Object |       |           |            |             | Subject |       |           |            |             |
|-------|--------|-------|-----------|------------|-------------|---------|-------|-----------|------------|-------------|
| File  | Rand.  | MFS   | word2word | word2class | class2class | Rand.   | MFS   | word2word | word2class | class2class |
| br-a01| .286   | .746  | .138      | .447       | .542        | .313    | .884  | .312      | .640       | .749        |
| br-b20| .233   | .776  | .093      | .418       | .487        | .292    | .780  | .354      | .580       | .677        |
| br-j09| .254   | .645  | .071      | .429       | .399        | .256    | .761  | .200      | .500       | .499        |
| br-r05| .269   | .639  | .126      | .394       | .577        | .294    | .720  | .144      | .601       | .710        |

**Table 3**. Average recall for the nouns in the four Semcor files.

Table 1 shows the data for the set of nouns. Note that only 19% (15%) of the occurrences of the nouns are objects (subjects) of any verb. Table 2 shows the average results using subject and object relations for each possible formalization. Each column shows respectively, the precision, the coverage over the occurrences with the given relation, and the recall. Random and most frequent baselines are also shown. Word-to-word gets the highest precision of all three, but it can only be applied on a few instances. Word-to-class gets slightly better precision than class-to-class, but class-to-class is near complete coverage and thus gets the best recall of all three. All are well above the random baseline, but slightly below the most frequent sense.

On the all-nouns experiment, we disambiguated the nouns appearing in four files extracted from Semcor. We observed that not many nouns were related to a verb as object or subject (e.g. in the file br-a01 only 40% (16%) of the polisemous nouns were tagged as object (subject)). Table 3 illustrates the results on this task. Again, word-to-word obtains the best precision in all cases, but because of the lack of data the recall is low. Class-to-class attains the best recall.

We think that given the small corpus available, the results are good. Note that there is no smoothing or cut-off value involved, and some decisions are taken with very little points of data. Sure enough both smoothing and cut-off values will allow to improve the precision. On the contrary, literature has shown that the most frequent sense baseline needs less training data.

## 7 Analysis of the acquired selectional preferences

In order to analyze the acquired selectional preferences, we wanted a word that did not occur too often and which had clearly

*Sense 1*
*church, Christian church, Christianity*
    => *religion, faith*
      => *institution, establishment*
        => *organization, organisation*
          => *social group*
            => *group, grouping*

*Sense 2*
*church, church building*
    => *place of worship, house of prayer,*
      *house of God, house of worship*
      => *building, edifice*
        => *structure, construction*
          => *artifact, artefact*
            => *object, physical object*
              => *entity, something*

*Sense 3*
*church service, church*
    => *service, religious service, divine service*
      => *religious ceremony, religious ritual*
        => *ceremony*
          => *activity*
            => *act, human action, human activity*

**Figure 1**. Word senses and superclasses for church

distinguishable senses. The goal is to study the preferences that were applied in the disambiguation for all occurrences, and check what is the difference among each of the models.

The selected word was *church*, which has three senses in WordNet, and occurs 19 times. Figure 1 shows the three word senses and the corresponding subsuming concepts. Table 4 shows the results of the disambiguation algorithm for church.

|              | #occ | OK | KO | No ansr | Prec. | Cov. | Rec. |
|---|---|---|---|---|---|---|---|
| obj MFS | 19 | 4 | 15 | 0 | .210 | 1.00 | .210 |
| obj word-to-word | 19 | 0 | 0 | 19 | .000 | .000 | .000 |
| obj word-to-class | 19 | 12 | 5 | 2 | .705 | .894 | .631 |
| obj class-to-class | 19 | 12 | 7 | 0 | .631 | 1.00 | .631 |
| subj MFS | 10 | 8 | 2 | 0 | .800 | 1.00 | .800 |
| subj word-to-word | 10 | 0 | 0 | 10 | .000 | .000 | .000 |
| subj word-to-class | 10 | 4 | 3 | 3 | .571 | .700 | .400 |
| subj class-to-class | 10 | 6 | 4 | 0 | .600 | 1.00 | .600 |

**Table 4:** Results disambiguating the word *church*.

In the word-to-word model, the model is unable to tag any of the examples[4] (all the verbs related to "church" were different). For church as object, both class-to-class and word-to-class have similar recall, but word-to-class has better precision. Notice that the majority of the examples with church as object were not tagged with the most frequent sense in Semcor, and therefore the MFS precision is remarkably low (21%). For church as subject, the class-to-class model has both better precision and coverage.

We will now study in more detail each of the examples.

### 7.1 Church as object

There were 19 examples with *church* as object (15 tagged in Semcor with sense 2 and 4 with sense 1). Using the word-to-class model, 12 were tagged correctly, 5 incorrectly and 2 had not enough data to answer. In the class-to-class model 12 examples were tagged correctly and 7 incorrectly. Therefore there was no gain in recall.

First, we will analyze the results of the **word-to-class model**. From the 12 hits, 10 corresponded to sense 2 and the other 2 to sense 1. Here we show the 12 verbs and the superconcept of the senses of church that gets the highest selectional preference probability, and thus selects the winning sense, in this case, correctly.

- Tagged with sense 2:
  look:     <building, edifice>
  have:     <building, edifice>

---

[4] Note that we applied 10fold crossvalidation. The model is not able to tag anything because the verbs in the testing samples do not appear in the training samples. In fact all the verbs governing church occur only once.

  demolish:   <building, edifice>
  move:       <structure, construction>
  support:    <structure, construction>
  build:      <structure, construction>
  enter:      <structure, construction>
  sell:       <artifact, artefact>
  abandon:    <artifact, artefact>
  see:        <artifact, artefact>
- Tagged with sense 1
  strengthen: <organization, organisation>
  turn_to:    <organization, organisation>

The five examples where the model failed revealed different types of errors. We will check each of the verbs in turn.

**1.** *Attend (Semcor 2, word-to-class 1)*[5]: We quote the whole sentence:

> From many sides come remarks that Protestant churches are badly **attended** and the large medieval cathedrals look all but empty during services .

We think that the correct sense should be 3 ("*church services*" are attended, not the buildings). In any case, the class that gets the higher weight is <*institution, establishment*>, pointing to sense 1 of *church* and beating the more appropriate class <*religious ceremony, religious ritual*> because of the lack of examples in the training.

**2.** *Join (Semcor 1, word-to-class 2)*: It seems that this verb should be a good clue for sense 1. But among the few occurrences of *join* in the training set there were "*join-obj-temple*" and "*join-obj-synagogue*". Both *temple* and *synagogue* have do not have organization-related concepts in WordNet and they were thus tagged with a concept under <*building,*

---

[5] For each verb we list the sense in Semcor (the correct reference sense) and the sense assigned by the model.

*edifice>*. This implies that *<place of worship, house of prayer, house of God, house of worship>* gets most credit and the answer is sense 2.

**3.** *Imprison (Semcor 1, word-to-class 3)*: The scarcity of training examples is very evident here. There are only 2 examples of *imprison* with an object, one of them wrongly selected by Minipar (*imprison-obj-trip*) that falls under *<act, human action, human activity>* and points to sense 3.

**4.** *Empty (Semcor 2, word-to-class 1)*: The different senses of *empty* introduce misleading examples. The best credit is given to *<group, grouping>* (following an sense of *empty* which is not appropriate here) which selects the sense 1 of church. The correct sense of *empty* in this context relates with *<object, physical object>*, and would thus select the correct sense, but does not have enough credit.

**5.** *Advance (Semcor 2, word-to-class 3)*: the misleading senses of "advance" and the low number of examples point to sense 3.

We thus identified 4 sources of error in the word-to-class model:
A. Incorrect Semcor tag
B. Wrongly extracted verb-object relations
C. Scarcity of data
D. Misleading verb senses

The **class-to-class** model should help to mitigate the effects of errors type C and D. We would specially hope for the class-to-class model to discard misleading verb senses. We now turn to analyze the results of this model.

From the 12 correct examples tagged using word-to-class, we observed that 3 were mistagged using class-to-class. The reason was that the class-to-class introduces new examples from verbs that are superclasses of the target verb, and these introduced noise. For example, we examined the verb *turn_to* (tagged in Semcor with sense 1):

**1.** *turn-to (Semcor 1, word-to-class 1)*: there are fewer training examples than in the class-to-class model and they get more credit. The relation "*turn_to-obj-platoon*" gives weight to the class *<organization, organisation>*.

**2.** *turn-to (Semcor 1, class-to-class 2)*: the relations "*take_up-obj-position*" and "*call_on-obj-esprit_de_corps*" introduce noise and point to the class *<artifact, artefact>*. As a result, the sense 2 is wrongly selected.

From the 5 mistagged examples in class-to-class, only "empty" was tagged correctly using classes (in this case the class-to-class model is able to select the correct sense of the verb, discarding the misleading senses of empty):

**1.** *Attend, Join, Advance*: they had errors of type A and B (incorrect Semcor tag/ misleading verb-object relations) and we can not expect the "class-to-class" model to handle them.

**2.** *Imprison*: still has not enough information to make a good choice.

**3.** *Empty (Semcor 2, class-to-class 2)*: new examples associated to the appropriate sense of *empty* give credit to the classes *<place of worship, house of prayer, house of God, house of worship>* and *<church, church building>*. With the weight of these classes the correct sense 2 is correctly chosen.

Finally, the 2 examples that received no answer in the "word-to-class" model were tagged correctly:

**1.** *Flurry (Semcor 2, class-to-class 2)*: the answer is correct although the choice is made with few data. The strongest class is *<structure, construction>*.

**2.** *Rebuild (Semcor 2, class-to-class 2)*: the new information points to the appropriate sense.

## 7.2 Church as subject

The class2class model showed a better behavior with the examples in which *church* appeared as subject. There were only 10 examples, 8 tagged with sense 1 and 2 with sense 2.

In this case, the class-to-class model tagged in the same way the examples tagged by the class-to-word model, but it also tagged the 3 occurrences that had not been tagged by the word-to-class model (2 correctly and 1 incorrectly).

## 8 Conclusions

We presented a statistical model that extends selectional preference to classes of verbs, yielding a relation between classes in a hierarchy, as opposed to a relation between a word and a class. The motivation is twofold: different senses of a verb may have different preferences, and some classes of verbs can share preferences.

The model is trained using subject-verb and object-verb relations extracted from a sense-disambiguated corpus using Minipar. A peculiarity of this exercise is the use of a small sense-disambiguated corpus, in contrast to using a large corpus of ambiguous words.

Contrary to other methods we do not try to find the classes which generalize best the selectional preferences. All possibilities, even the ones with very low probability, are stored.

Evaluation is based on a word sense disambiguation exercise for a sample of words and a sample of documents from Semcor. The proposed model gets similar results on precision but significantly better recall than the classical word-to-class model.

We plan to train the model on a large untagged corpus, in order to compare the quality of the acquired selectional preferences with those extracted from this small tagged corpora. The model can easily be extended to disambiguate other relations and POS. At present we are also integrating the model on a supervised WSD algorithm that uses decision lists.